\documentclass{scrartcl}
\usepackage[top=2.5cm, bottom=3cm,footskip=8mm, left=2cm, right=2cm]{geometry}
\KOMAoption{captions}{tableheading}
\usepackage{subcaption}
\usepackage{siunitx}
\usepackage{amsmath}
\usepackage{hyperref}
\usepackage{graphicx} 
\usepackage{xcolor}
\xdefinecolor{darkblue}{rgb}{0.0 0.0 0.4}
\graphicspath{{./Graphics}}
\usepackage{authblk}
\usepackage{booktabs}
\usepackage{cleveref}
\usepackage[abbreviations,symbols,nogroupskip,nonumberlist]{glossaries-extra}

\makeglossaries[abbreviations]
\newabbreviation{ai}{AI}{artificial intelligence}
\newabbreviation{flops}{FLOPs}{floating-point operations}
\newabbreviation{kpi}{KPI}{key performance indicator}
\newabbreviation{sbc}{SBC}{singleboard computer}
\newabbreviation{ram}{RAM}{random-access memory}
\newabbreviation{rom}{ROM}{read-only memory}
\newabbreviation{diy}{DIY}{do it yourself}
\newabbreviation{dsp}{DSP}{digital signal processing} 
\newabbreviation{gpio}{GPIO}{general-purpose input/output}
\newabbreviation{iot}{IoT}{internet of things}
\newabbreviation{ppk}{PPK}{Power Profiler Kit} 
\newabbreviation{onnx}{ONNX}{Open Neural Network Exchange} 
\newabbreviation{atp}{ATP}{access all the pins}

\providecommand{\mytitle}{Pareto Optimal Benchmarking of AI Models on ARM Cortex Processors for Sustainable Embedded Systems}
\hypersetup{%
    colorlinks=true, 
    hypertexnames=true, pdfhighlight=/O,
    linkcolor=darkblue, urlcolor=black, citecolor=darkblue, 
    pdftitle={Pranay Jain et al.: \mytitle},%
    pdfauthor={\textcopyright\ Pranay Jain},%
    pdfsubject={},%
    pdfkeywords={},%
    pdfcreator={pdfLaTeX},%
    pdfproducer={LaTeX}%
}
\title{\mytitle}
\author[1,2]{Pranay Jain}
\author[1]{Maximilian Kasper}
\author[2]{Göran Köber}
\author[2,3]{Oliver Amft}
\author[1]{Axel Plinge}
\author[1,4]{Dominik Seuß}

\affil[1]{Fraunhofer Institute for Integrated Circuits IIS, Germany}
\affil[2]{Intelligent Embedded Systems (IES) - Lab, University of Freiburg,
Germany}
\affil[3]{Hahn-Schickard, Germany}
\affil[4]{Center for Artificial Intelligence and Robotics (CAIRO),
Technische Hochschule Würzburg-Schweinfurt, Germany}

\date{}

\begin{document}
\maketitle
\begin{abstract}
    This work presents a practical benchmarking framework for optimizing \gls{ai}
    models on ARM Cortex processors (M0+, M4, M7), focusing on energy
    efficiency, accuracy, and resource utilization in embedded systems. Through
    the design of an automated test bench, we provide a systematic approach
    to evaluate across \glspl{kpi} and identify optimal
    combinations of processor and \gls{ai} model. The research highlights a nearlinear
    correlation between \gls{flops} and inference
    time, offering a reliable metric for estimating computational demands. Using
    Pareto analysis, we demonstrate how to balance trade-offs between energy
    consumption and model accuracy, ensuring that \gls{ai} applications meet performance
    requirements without compromising sustainability. Key findings
    indicate that the M7 processor is ideal for short inference cycles, while the
    M4 processor offers better energy efficiency for longer inference tasks. The
    M0+ processor, while less efficient for complex \gls{ai} models, remains suitable
    for simpler tasks. This work provides insights for developers, guiding them
    to design energy-efficient \gls{ai} systems that deliver high performance in realworld
    applications.
\end{abstract}


\glsresetall

\section{Introduction}
The integration of \gls{ai} into embedded systems is challenged
by the need to balance model performance with energy consumption,
a crucial factor for the sustainability and practicality of these systems. Edge
\gls{ai} provides energy efficiency, low latency, and privacy directly on the
device, making them critical for applications from smart home devices to
autonomous vehicles~\cite{jacob2018quantization, deutel2025combining}.

\begin{table}
    \centering    
    \begin{tabular}{lccccrc}
        \toprule
        \textbf{Platform} &
        \textbf{Freq.} &
        \textbf{Memory} &
        \textbf{Storage} &
        \textbf{Power} &
        \textbf{Price} &
        \textbf{CO2 Footprint} 
        \\ \midrule \midrule
        
        Cloud & GHz & $>10$\qty{}{\giga\byte} & TBs-PBs & $\approx$\qty{1}{\kilo\watt} & $>1000\$$ & Hundreds of kgs \\ \midrule
        Mobile & GHz & Few GBs & GBs & $\approx$\qty{1}{\watt} & $>100\$$ & Tens of kgs \\ \midrule
        Edge \gls{ai} & MHz & KBs & Few MBs & $\approx$\qty{1}{\milli\watt} & $>10\$$ & Single kgs \\
        \bottomrule
    \end{tabular}
    \caption{Cloud vs. mobile vs. edge \gls{ai} systems across different parameters}
    \label{tbl:cloud_vs_edge}
\end{table}

As shown in Table~\ref{tbl:cloud_vs_edge}, edge \gls{ai} platforms operate under strict resource
constraints, with significantly less memory, power, and a smaller CO2 footprint
compared to their cloud and mobile counterparts~\cite{prakash2301tinyml}. While considerable
research has been conducted on benchmarking \gls{ai} performance on \glspl{sbc}
like Raspberry Pi, there is a gap in studies focusing
on bare-metal processors. \glspl{sbc}, while popular and accessible, introduce
additional layers of abstraction that can obscure the true performance characteristics
of the hardware. In contrast, bare-metal processors provide a more
direct and granular understanding of the hardware capabilities, free from
the overhead introduced by operating systems and middleware. This study
addresses the need for a dedicated test bench that evaluates embedded \gls{ai}
systems at the bare-metal level. \glspl{kpi} such as accuracy,
inference time, and energy consumption are measured, and Pareto front
analysis is applied to identify optimal trade-offs. The results provide a solid
basis for selecting \gls{ai} models that achieve high efficiency even under strict
resource constraints.

The rest of the paper is structured as follows. \Cref{sec:related_work} reviews related
work and highlights the need for standardized benchmarking in embedded \gls{ai}.
\Cref{sec:methodology} outlines the methodology, including the design of the benchmarking
framework, model selection, and experimental setup. \Cref{sec:results} presents the 
benchmarking results, with a focus on trade-offs between accuracy, latency,
and energy efficiency. Finally,~\cref{sec:discussion} and~\ref{sec:conclusion} suggest directions for future
research and conclude the paper.

\section{Related Work}
\label{sec:related_work}
This work is positioned at the intersection of sustainability in edge \gls{ai} systems
and the benchmarking of \gls{ai} models on embedded platforms. Existing
literature provides valuable insights into these domains, yet notable gaps
remain that motivate the contributions of this study.

The sustainability of Edge \gls{ai} has gained attention as the growth of
\gls{iot} devices contributes to rising carbon emissions and electronic waste~\cite{prakash2301tinyml}.
Performing \gls{ai} inference directly on resource-constrained microcontrollers
offers significant reductions in energy use compared to cloud-based computation.
The life cycle assessment study by~\cite{prakash2301tinyml} highlights the potential for this
approach to lower emissions across diverse applications but also points out its
non-trivial footprint when scaled globally. Their findings underline the need
for energy-aware design principles that consider the environmental trade-offs
of deploying edge \gls{ai} at scale. However, fine-grained studies on bare-metal
processors remain limited.

Benchmarking \gls{ai} models on embedded systems has been widely studied,
with challenges including restricted memory, limited energy budgets, and
heterogeneous architectures~\cite{banbury2020benchmarking}. MLPerf Tiny~\cite{banbury2021mlperf} provides a standardized
suite for assessing the latency, accuracy, and energy efficiency of small-scale
models, ensuring reproducibility across platforms. DeepEdgeBench~\cite{baller2021deepedgebench}
expands this scope by evaluating multiple neural networks across accelerators
and processors, focusing on inference time, memory footprint, and energy
use. LwHBench~\cite{garcia2023analysing} further benchmarks lightweight models on \glspl{sbc},
analyzing inference speed, accuracy, and power consumption
to guide deployment decisions. These frameworks have advanced
evaluation practices for embedded \gls{ai}, but they concentrate largely on \glspl{sbc}
and higher-level platforms, where operating systems and middleware can
obscure hardware-specific behavior.

Model compression techniques particularly structured pruning~\cite{han2015deep} and
static quantization~\cite{jacob2018quantization} are widely used to reduce model size and energy
consumption for edge deployment. Meanwhile, multi-objective Bayesian
optimization has emerged as a powerful tool to navigate trade-offs between
accuracy, latency, and energy~\cite{deutel2025combining}. However, to the best of the authors
knowledge, these methods have not been systematically evaluated in baremetal settings.

\section{Methodology}
\label{sec:methodology}
\begin{figure}
    \centering
    \includegraphics[width=\textwidth]{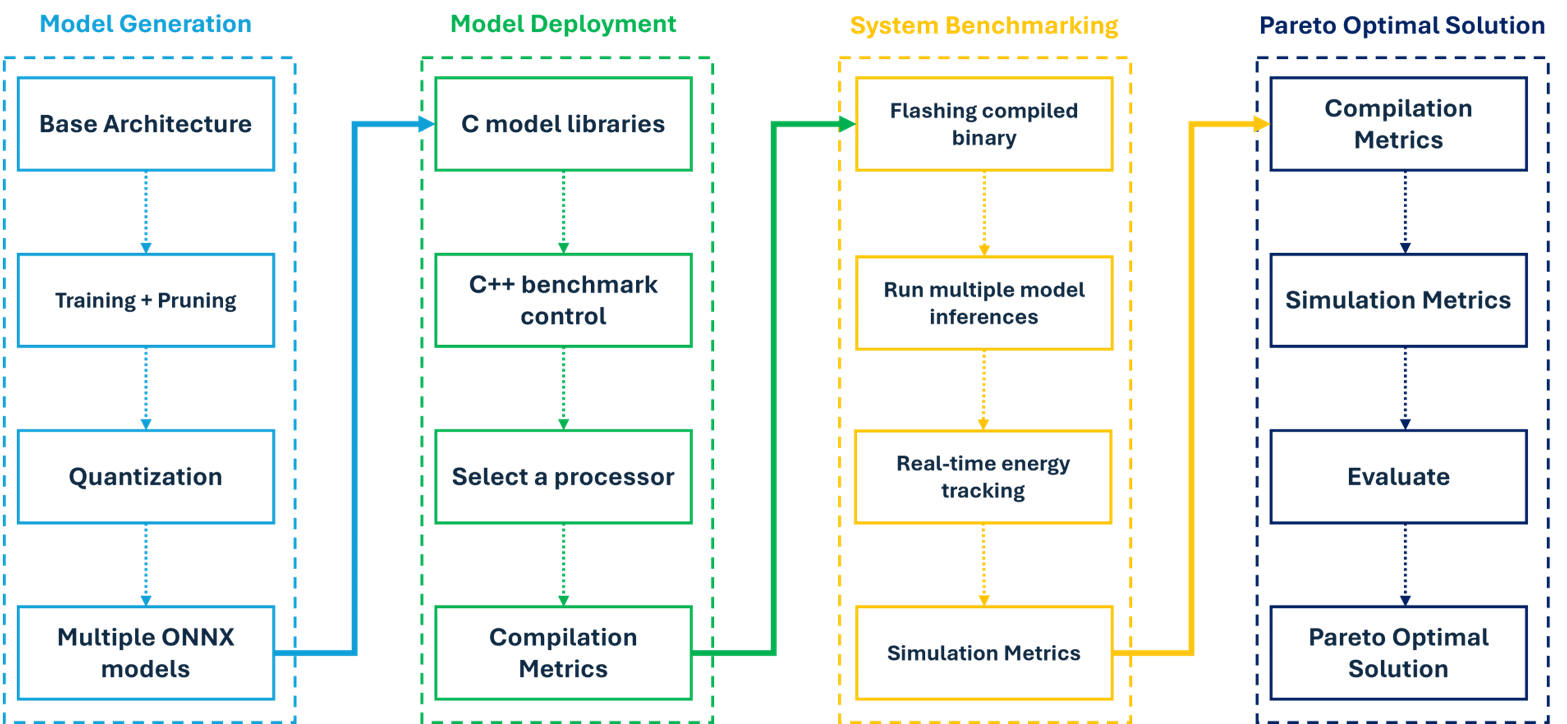}
    \caption{Overview of the test bench architecture and workflow.}
    \label{fig:overview}
\end{figure}

The proposed test bench is depicted in Figure~\ref{fig:overview} and follows a structured
workflow that systematically evaluates \gls{ai} models on bare-metal processors
to achieve energy-efficient embedded \gls{ai} design. The workflow consists of
four key stages: Model Generation, Model Deployment, System Benchmarking,
and Pareto Optimal Solution, each contributing to the comprehensive
assessment of \gls{ai} models in resource-constrained environments.

The workflow begins in the Model Generation stage, starting with a
baseline architecture for a given use case. To explore various trade-offs
between performance and resource requirements, we employ an automated
multi-objective optimization method to guide the application of structured
pruning~\cite{han2015deep, deutel2025combining}. This process generates a diverse population of pruned models.
Following this, each model undergoes static 8-bit quantisation to further
reduce memory footprint and improve computational efficiency~\cite{jacob2018quantization}. The final
output of this stage is a collection of multiple models in the \gls{onnx} format,
each representing a unique trade-off point.

The Model Deployment stage converts the \gls{onnx} models into executable
binaries tailored for the target hardware. First, each \gls{onnx} model is translated
into a self-contained C model library. This library is then integrated with
a C++ benchmarking framework designed to control the inference execution 
and data collection. The test bench allows for the selection of a specific
target processor (\emph{Cortex-M0+, M4, or M7}), ensuring the code is compiled for
the correct architecture. Finally, the C++ control code and the C model are
compiled together, producing a single executable binary ready for the System
Benchmarking stage.

In the System Benchmarking stage, the compiled binary is flashed onto
the target processor. The benchmarking framework then executes multiple
model inferences to gather stable performance data. During this execution, we
use real-time energy tracking to precisely measure power consumption, while
simultaneously recording key inference metrics such as execution latency.

Finally, the Pareto Optimal Solution stage processes the benchmarking
results to determine the most efficient \gls{ai} model configuration by analyzing
compilation metrics (e.g. computational overhead) and inference metrics
(e.g., latency, accuracy, and power consumption), while also considering
use case and hardware specific factors such as idle time between inference
cycles, since idle power consumption can significantly affect overall efficiency.
Through this systematic evaluation, models are compared to identify
optimal trade-offs, ensuring that the final deployed \gls{ai} model achieves the best
balance between energy efficiency and performance for resource-constrained
embedded environments.

This structured workflow provides a systematic approach to benchmarking
\gls{ai} models on bare-metal processors, enabling precise evaluation and
optimization for energy-efficient Edge \gls{ai} applications.

\subsection{Use Cases}

\begin{table}
    \centering
    \begin{tabular}{p{3.7cm}p{3.3cm}rp{2.8cm}r}
        \toprule
         \textbf{Use case} &
         \textbf{Model} &
         \textbf{\#Params} &
         \textbf{Dataset} &
         \hspace{-3em}\textbf{Quality Target}
         \\ \midrule \midrule
         Image Classification & ResNet~\cite{banbury2021mlperf, he2016deep} & \textbf{78k} & CIFAR-10~\cite{cifar10} & $\ge \qty{80}{\percent}$ \\ \midrule
         Optical Digit Recognition & LeNet-5~\cite{lecun2002gradient} & \textbf{140k} & MNIST~\cite{lecun2002gradient} & $\ge \qty{95}{\percent}$ \\ \midrule
         Anomaly Detection & Autoencoder \cite{banbury2021mlperf} & \textbf{269k} & ToyADMOS \cite{koizumi2019toyadmos} & AUC $\ge \qty{85}{\percent}$ \\ \midrule
         Visual Wake Words & MobileNetV1 \cite{howard2017mobilenets} & \textbf{3198k} & MSCOCO14 \cite{lin2014microsoft} & $\ge \qty{80}{\percent}$ \\
         \bottomrule
    \end{tabular}
    \caption{Overview of benchmark \gls{ai} use-cases}
    \label{tab:use_case_overview}
\end{table}
The selected use-cases are representing diverse application domains relevant
to embedded \gls{ai} benchmarking and are adopted from~\cite{banbury2021mlperf}. 
Table \ref{tab:use_case_overview} provides an overview.
Optical Digit
Recognition was additionally included and serves as a foundational computer
vision task, leveraging LeNet-5 on MNIST~\cite{lecun2002gradient}. We set a target accuracy
of \qty{95}{\percent}, as MNIST is largely considered a solved task but remains a useful
benchmark for assessing high-accuracy performance under edge constraints.
Anomaly Detection employs an autoencoder trained on industrial sound
datasets~\cite{banbury2021mlperf, koizumi2019toyadmos}, targeting robust detection when only normal data is
available. Compact Image Classification relies on a customized ResNet architecture~\cite{banbury2021mlperf, he2016deep} optimized for embedded constraints, evaluated on CIFAR-10~\cite{lecun2002gradient}. Finally, Visual Wake Words uses MobileNetV1~\cite{howard2017mobilenets} for binary person
detection on MSCOCO~\cite{lin2014microsoft}, reflecting \gls{iot}-oriented scenarios.

\subsection{Experimental Setup}
\begin{figure}
    \centering
    \includegraphics[width=\textwidth]{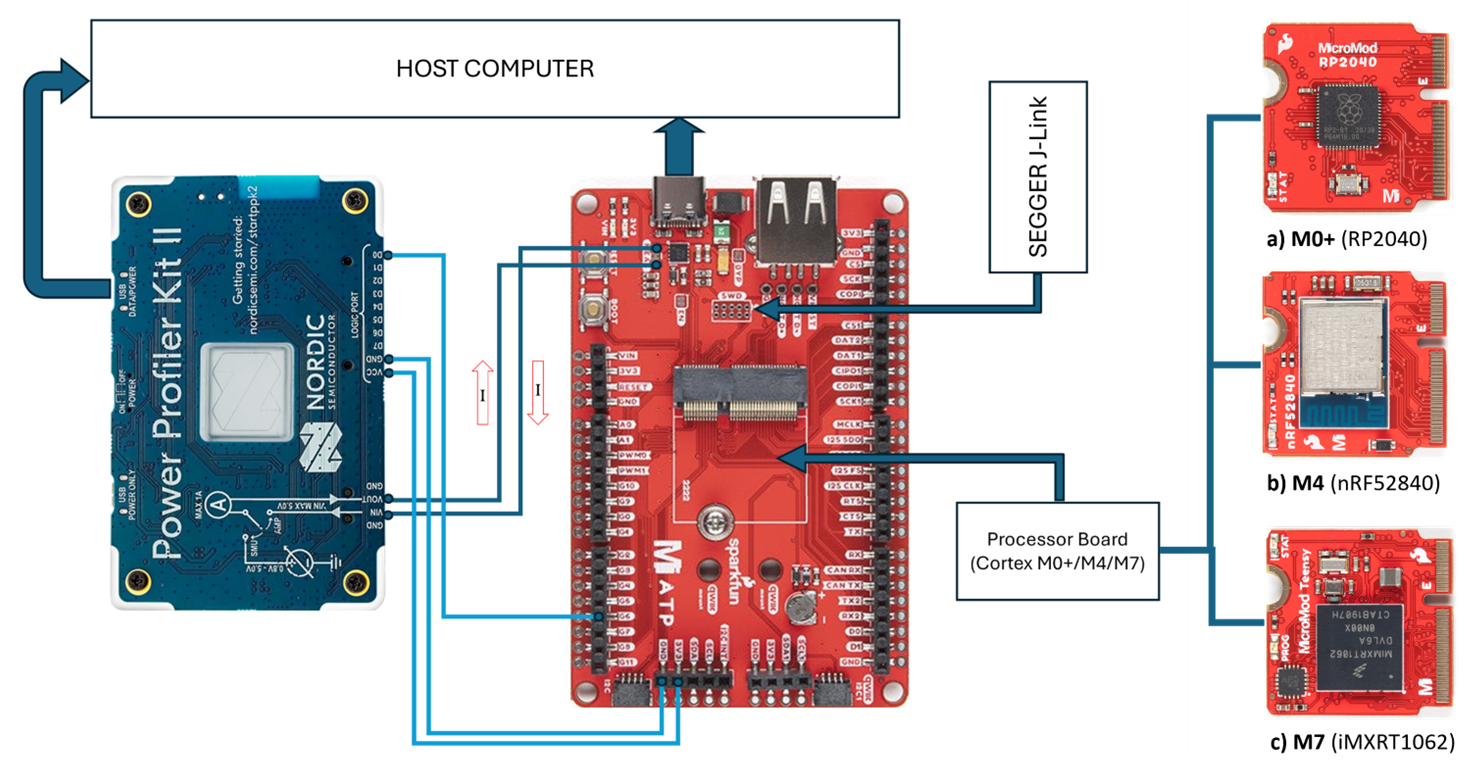}
    \caption{Experimental setup for test bench evaluation.}
    \label{fig:experimental_setup}
\end{figure}

The experimental setup is designed to systematically evaluate the performance
and energy efficiency of \gls{ai} models deployed on bare-metal processors.
As illustrated in Figure~\ref{fig:experimental_setup}, the setup consists of key hardware
components, including an \gls{atp} carrier board, a Segger J-Link debugger, a \gls{ppk}, and associated signal connections to facilitate
accurate measurement of inference-related metrics.

The \gls{atp} carrier board serves as the primary hardware platform for
executing \gls{ai} models across different embedded processors. It supports multiple
processor configurations, including M0+, M4, and M7 cores, enabling
comparative benchmarking of \gls{ai} models under varying computational capabilities.
The deployment process involves flashing the \emph{model.hex} file onto the respective processor.
For the M0+ and M4 processors, the Segger J-Link
debugger is used for programming, ensuring precise execution and debugging
capabilities. In contrast, the M7 processor is flashed via the USB-C interface,
providing a streamlined deployment process.

\begin{figure}
    \centering
    \includegraphics[width=\textwidth]{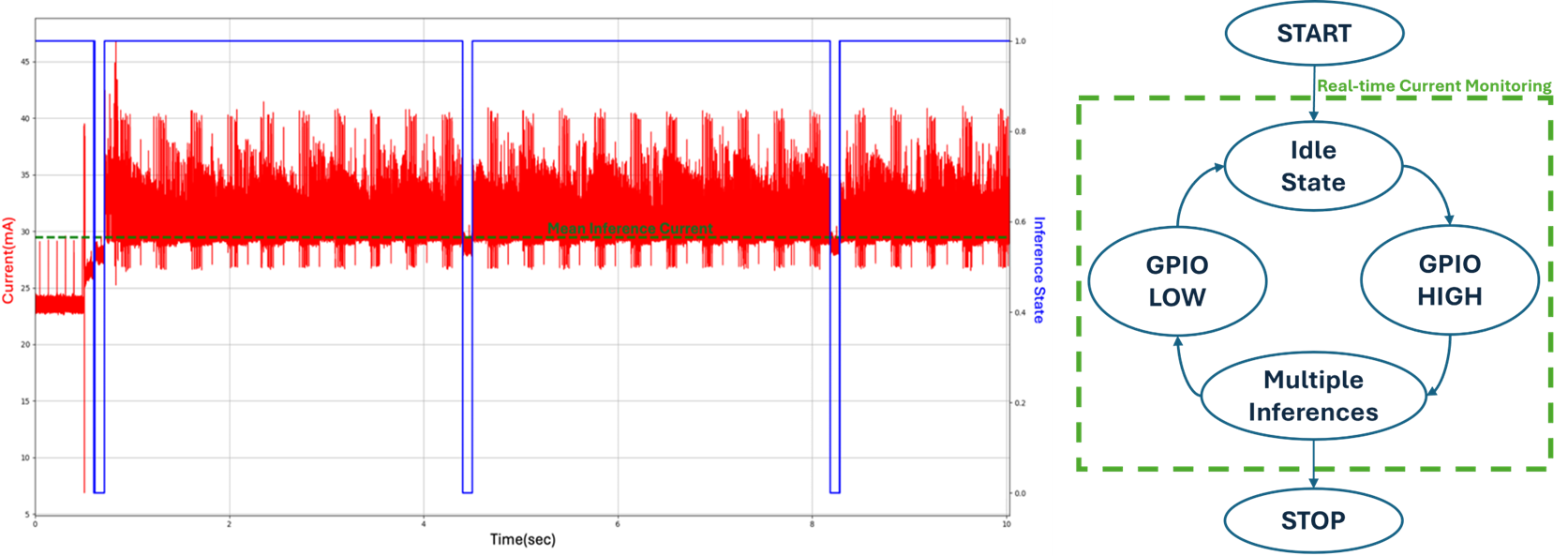}
    \caption{Benchmarking flow diagram (left) and exemplary current measurement (right).}
    \label{fig:benchmarking_flow}
\end{figure}

To measure energy consumption during model inference, a \gls{ppk} is integrated into the setup.
The \gls{ppk}’s \emph{Vin} and \emph{Vout} pins are connected to the \emph{MEAS} pins on the \gls{atp} carrier board, enabling real-time
measurement of the current flowing through the board. This configuration
provides critical insights into the power and energy consumption across
different \gls{ai} models and hardware configurations. Additionally, to accurately
mark the inference execution phases, a \gls{gpio} pin on the \gls{atp} carrier board is
connected to the \emph{D0} pin of the \gls{ppk}. This connection allows the \gls{ppk}’s state to
be toggled as shown in Figure~\ref{fig:benchmarking_flow}, ensuring precise synchronization of power
measurements with inference execution.

The flow diagram on the left in Figure \ref{fig:benchmarking_flow} illustrates the control sequence
for toggling a \gls{gpio} pin and executing multiple inferences per active phase.
The plot on the right depicts the measured current consumption (red) and
inference state (blue) over time, with the dashed green line indicating
the mean current during active processing. Within each cycle, repeated
inference execution ensures sufficient measurement duration and enables
averaging of current values across multiple runs, resulting in more reliable
and representative measurements.

This experimental setup ensures a robust and reliable evaluation of \gls{ai}
model efficiency on embedded processors by facilitating direct power measurements
and enabling comparative performance analysis across different
core architectures.

\section{Results}
\label{sec:results}
This chapter presents the key results of the study, including test-bench reliability,
model variation, the relationship between \gls{flops} and inference time,
and a Pareto analysis of varying inference cycle times to reveal trade-offs
between energy consumption and model accuracy.

\subsection{Test-bench Reliability}
To validate the reliability of our test bench, each \gls{ai} model was benchmarked
five times to assess the consistency of the measurements. The variance across
key metrics like inference current, time, and especially energy was minimal.
This negligible variability confirms the stability and robustness of our
experimental setup, ensuring the trustworthiness of the subsequent findings.

\subsection{Model Size across Use Cases}
\begin{figure}
    \centering
    \includegraphics[width=\textwidth]{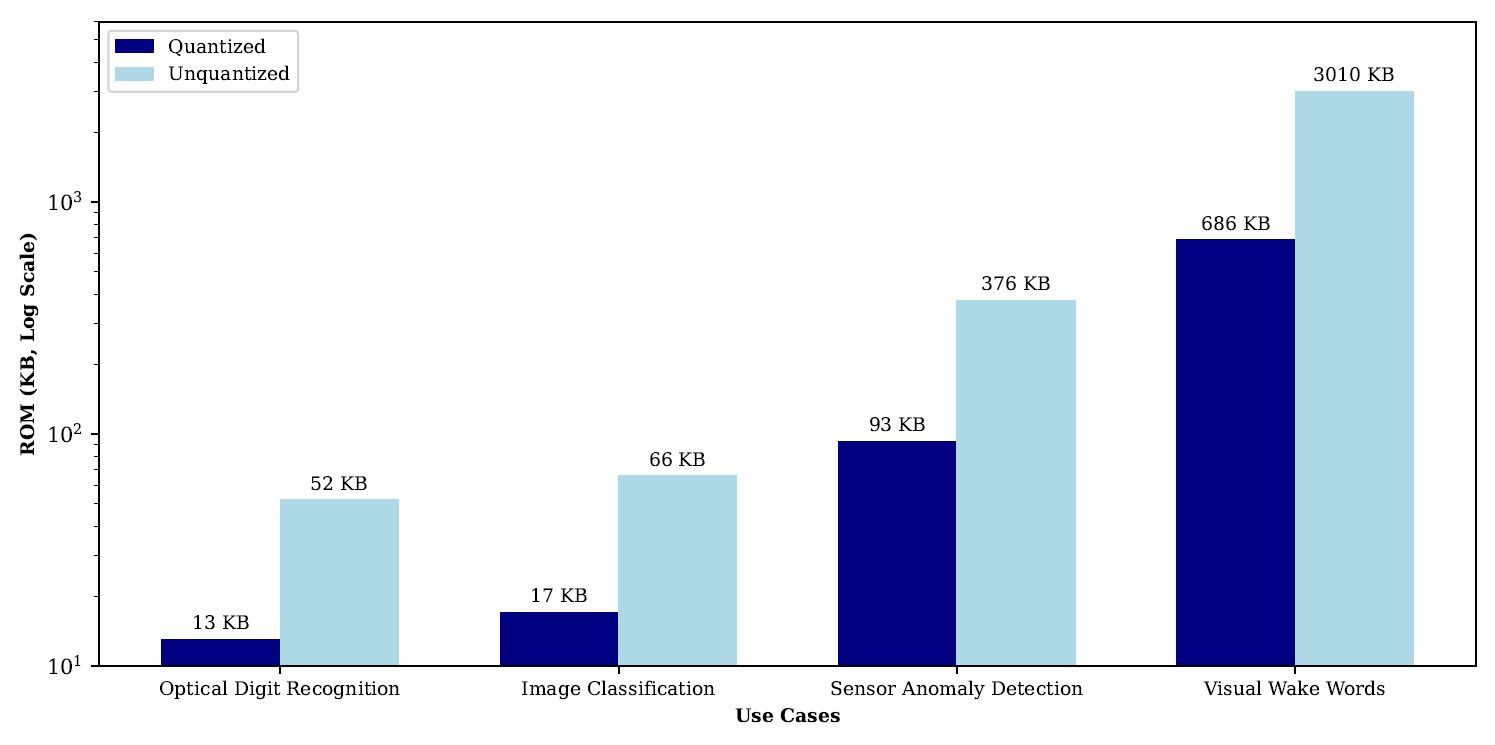}
    \caption{Mean \gls{ai} model sizes per use-case.}
    \label{fig:model_size}
\end{figure}

Figure~\ref{fig:model_size} shows the \gls{rom} requirements of quantized and unquantized
models across the benchmarked use cases (log scale). Quantization yields
a substantial reduction in model size, often to one-quarter of the original,
making deployment on constrained devices feasible. Among the tasks, \emph{Optical
Digit Recognition} is the smallest, while \emph{Visual Wake Words} is nearly
$50\times$, highlighting the wide range of memory footprints. Notably,
unquantized Visual Wake Words models exceeded the available memory on
the target processors.

\begin{figure}
    \centering
    \includegraphics[width=\textwidth]{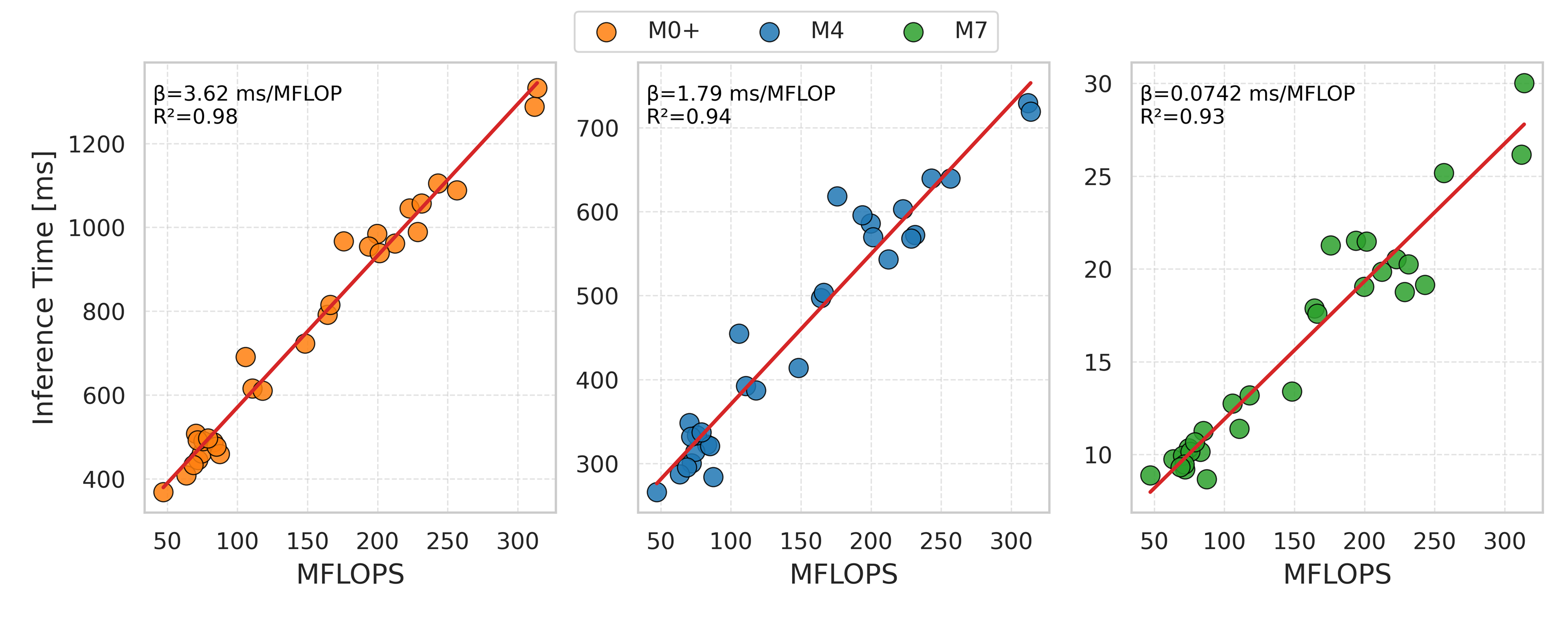}
    \caption{Linear dependency between inference time and \gls{flops}.}
    \label{fig:linear_reg}
\end{figure}

Through our experiments, \gls{ram} and \gls{rom} usage were found to be poor predictors
of energy consumption and are therefore considered primarily as hardware
constraints that determine whether a model can be deployed on a given
device. In contrast, Figure~\ref{fig:linear_reg} shows a linear relationship between \gls{flops}
and inference time. This correlation indicates that models with higher \gls{flops}
tend to have longer inference times, highlighting the importance of \gls{flops} as
a critical metric in evaluating model efficiency. Understanding this relationship
helps to anticipate the computational demands of the models and make
informed decisions about resource allocation and optimization strategies.

The analysis of the relationship between \gls{flops} and inference time
for the M0+, M4, and M7 processors (Figure~\ref{fig:linear_reg}) reveals a strong linear
correlation (R$^2 \ge 0.93$). This indicates that \gls{flops} can serve as a reliable
predictor of inference time within the evaluated range. The results 
highlight that while all processors exhibit strong linear scaling,
their computational efficiencies differ substantially — the M0+ and M4 show
a higher sensitivity, whereas the M7 achieves significantly faster inference
per FLOP due to its more advanced architecture.

\begin{figure}
    \centering
    \includegraphics[width=\textwidth]{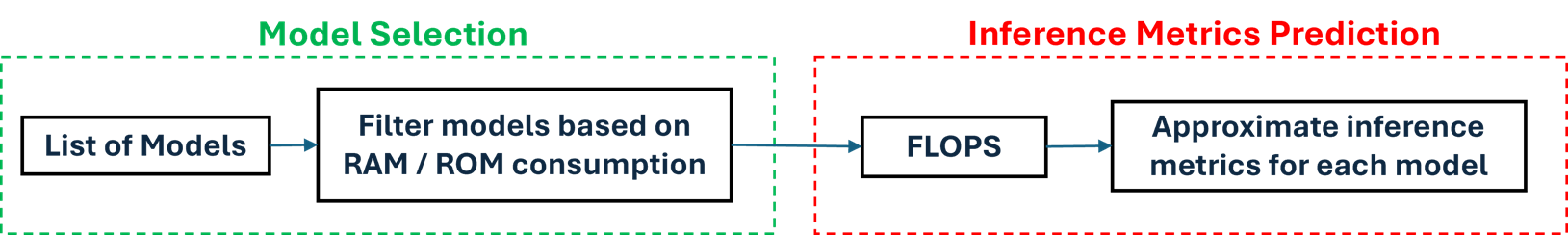}
    \caption{Process flow for model selection and inference metrics prediction.}
    \label{fig:process_flow}
\end{figure}

This established relationship supports using \gls{flops} as a predictive metric
for inference time on embedded platforms. As illustrated in Figure~\ref{fig:process_flow}, this
allows developers to estimate performance early in the design phase: once
model feasibility is confirmed via \gls{ram} and \gls{rom} constraints, \gls{flops} can
be used to approximate expected inference latency. This approach facilitates
processor–model co-design, enabling more efficient benchmarking and
selection in constrained environments.

For instance, if a particular processor’s \gls{ram} and \gls{rom} capacity align
with the requirements of a given model, its \gls{flops} can then be used to
approximate the expected inference time and inference energy. This benchmarking
analysis not only validates the use of \gls{flops} as a predictive metric
for inference time but also highlights the potential for using it in conjunction
with \gls{ram} and \gls{rom} specifications to guide the selection of the most
appropriate processor for specific applications.

\subsection{Analysis of Inference Cycle Energy}
Our energy efficiency analysis focuses on the total energy consumed per
inference cycle, a metric combining both the active inference period and the
subsequent idle time. The energy consumed during the idle state is especially
critical in applications with infrequent events, as it can dominate the total
power budget.

The processors exhibit significant differences in their datasheet deep sleep
currents. The Cortex-M4 is highly optimized for low-power states with an idle
current of just \qty{0.30}{\milli\ampere}, making it far more efficient during inactivity than the
M0+ (\qty{4.20}{\milli\ampere}) and the M7 (\qty{1.60}{\milli\ampere}). This idle current profoundly impacts
overall efficiency, as demonstrated when analysing the \glspl{kpi} of the full inference cycle.

\begin{figure}
    \centering
    \includegraphics[width=\textwidth]{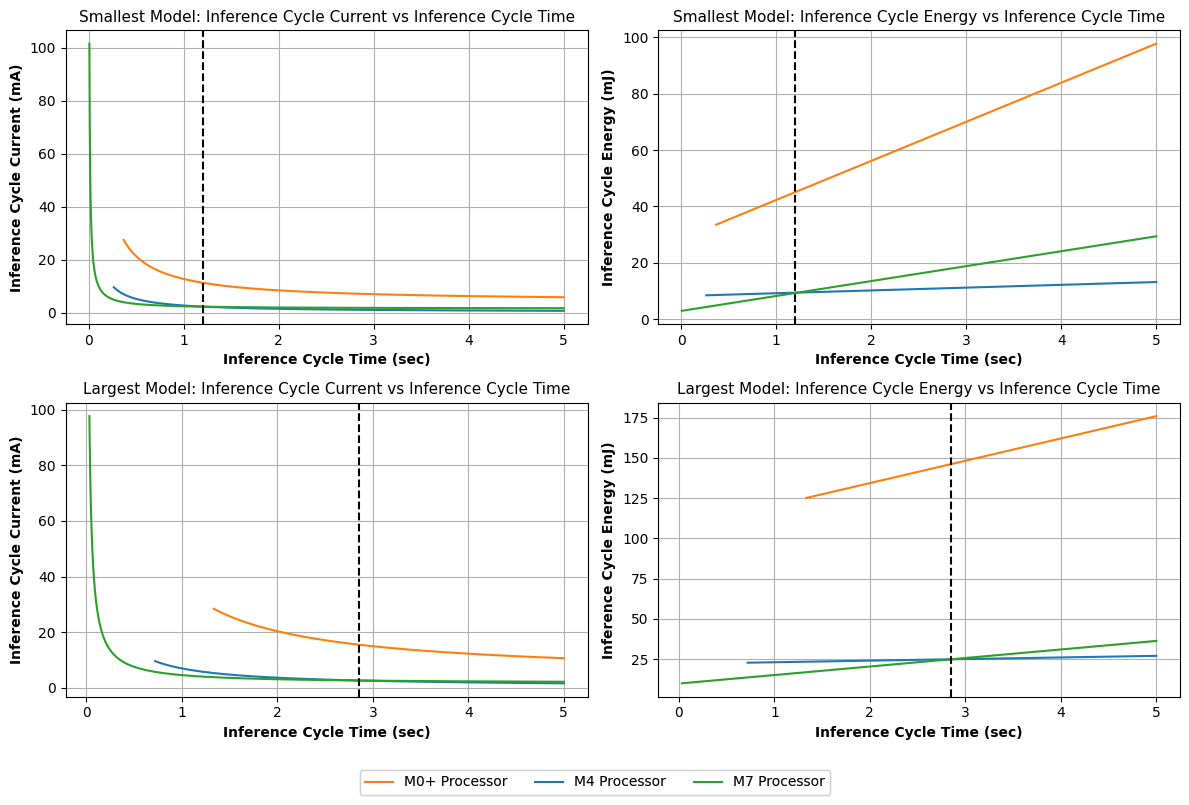}
    \caption{Inference cycle current and energy vs cycle time across processors for the smallest and largest models.}
    \label{fig:inference_cycle_current}
\end{figure}

Figure~\ref{fig:inference_cycle_current} shows the total inference cycle energy for each processor
over varying cycle times (0 - \qty{5}{\second}), comparing the smallest and largest models
obtained from our optimization to illustrate the performance range. Because
the supply voltage remains constant at \qty{3.3}{\volt}, power trends directly reflect
current measurements.

The data reveal distinct energy–performance characteristics. The M0+
processor consistently exhibits the highest energy consumption due to its
elevated inference and idle currents. In contrast, the M4 processor becomes
increasingly efficient at longer cycle times, where its low idle current
dominates the energy budget. The M7 processor, benefiting from its high
computational speed, achieves the lowest energy consumption during short
inference cycles by minimizing active power-on duration.

These results emphasize the trade-offs between processor architectures:
the M7 is best suited for short, frequent inference tasks, while the M4 offers
superior efficiency for longer, intermittent workloads. The M0+ remains the
least efficient across all scenarios. 

\subsection{Analysis of Energy Efficiency and Accuracy Trade-offs}
\begin{figure}
    \centering
    \includegraphics[width=\textwidth]{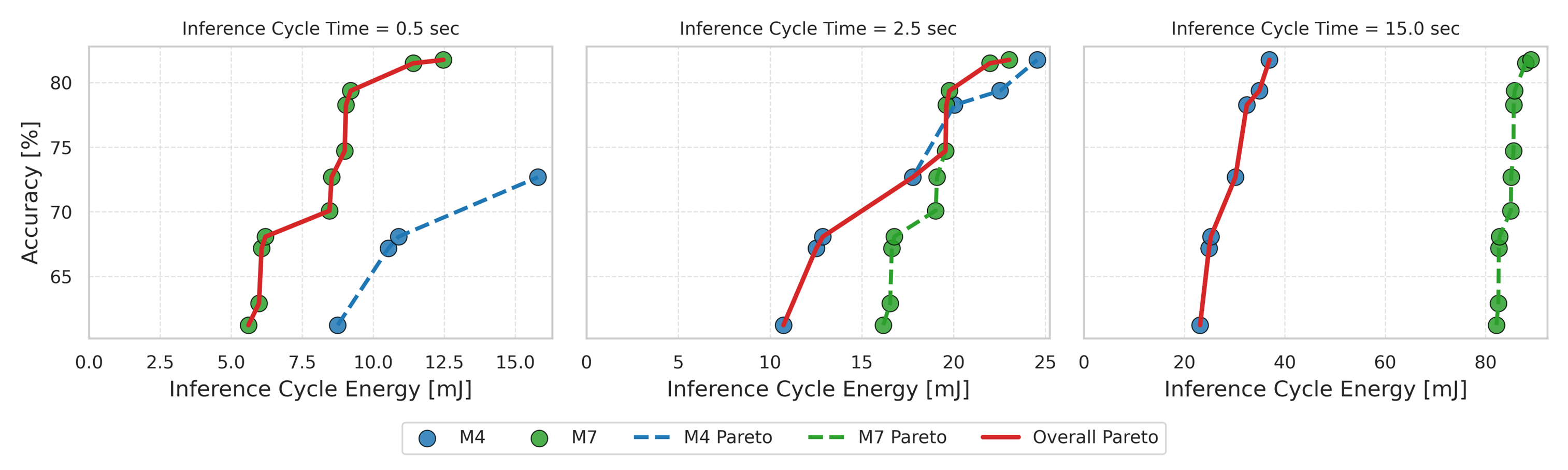}
    \caption{Energy vs. accuracy trade-offs for M4 and M7 across short, medium, and long
inference cycles. Pareto fronts highlight optimal choices per workload.}
    \label{fig:energy_acccuracy}
\end{figure}
Given its consistently higher inference and idle currents, the M0+ processor
was excluded from further energy–accuracy analysis, as it offers no competitive
advantage under the evaluated constraints. Figure~\ref{fig:energy_acccuracy} presents a Pareto
front analysis comparing inference cycle energy and accuracy across varying
cycle times (\qty{0.5}{\second}, \qty{2.5}{\second}, and \qty{5.0}{\second})
for the M4 and M7 processors.

The results reveal a fundamental shift in design priorities depending on
how frequently inference occurs. When inference tasks are frequent — such as
in applications with short cycle times (e.g., $\le$ \qty{0.5}{\second}) — the energy consumed
during active computation dominates the total budget. In this regime, the
Cortex-M7’s high computational throughput allows it to complete inference
significantly faster, resulting in lower total energy for a given level of
accuracy. Here, model efficiency directly translates into system-level savings,
and even modest reductions in inference time can yield meaningful
energy benefits. Consequently, models should be optimized primarily for
speed, with accuracy maintained only to the extent necessary for functional
correctness.

In stark contrast, when inference is infrequent and long idle periods
separate executions-as in applications with cycle times of \qty{2.5}{\second} or
more - the energy consumed while waiting far outweighs that used during computation.
Under these conditions, the Cortex-M4’s ultra-low idle current becomes decisive,
making it the more energy-efficient choice despite its slower inference
speed. More importantly, in this idle-dominated regime, the marginal energy
cost of running a slightly slower but more accurate model becomes negligible.
The execution time contributes little to the total energy budget, so there is
little to gain from aggressive model compression or latency optimization.
Instead, designers should favour models that maximize accuracy, as the
performance benefit during the rare inference event outweighs the minimal
energy penalty of a longer active phase.

This duality underscores a critical principle for embedded \gls{ai} deployment:
the optimal balance between energy and accuracy is not fixed - it is dictated
by the application’s temporal behaviour. Processor and model selection must
therefore be guided not only by hardware capabilities and model metrics, but
also by the expected duty cycle of the target use case. The Pareto fronts in
Figure~\ref{fig:energy_acccuracy} thus serve not just as performance benchmarks, but as a decision
map for co-designing hardware, models, and application timing.

\section{Discussion and Future Work}
\label{sec:discussion}
This study provides embedded \gls{ai} developers with a practical, applicationdriven
framework for selecting hardware and optimizing models — moving
beyond simplistic "faster is better” assumptions. Instead, the optimal choice
depends on how a processor’s energy profile aligns with the temporal
structure of the target application.

Inference cycle duration emerges as the primary determinant of architectural
preference. For applications demanding frequent, low-latency inference,
such as real-time industrial monitoring ($\le$ \qty{0.5}{\second} cycles), the Cortex-M7’s high
computational throughput dominates energy efficiency. Its \gls{dsp}-enhanced
architecture minimizes active execution time, making it ideal when inference
is the dominant energy sink. Conversely, for battery-powered devices with
long idle intervals, such as environmental sensors ($\ge$ \qty{5}{\second} cycles), the Cortex-
M4’s ultra-low idle current becomes decisive. Here, minimizing power
during waiting periods outweighs the benefit of faster computation. The
Cortex-M0+, while unsuitable for frequent-inference \gls{ai} workloads due to its
high idle consumption, retains value in non-\gls{ai} contexts and \gls{diy} projects,
where cost, accessibility, and development simplicity are prioritized over
performance.

Critically, this dichotomy informs not only processor selection, but also
model optimization strategy. In short-cycle regimes, where active computation
dominates energy use, reducing inference latency directly improves
system efficiency favouring compact, fast models. However, in long-cycle regimes
, the energy cost of inference becomes negligible compared to idle
consumption. Here, developers can prioritize accuracy over latency, selecting
higher-performing models without significant energy penalty.

To aid this complex decision-making process, our work establishes
two practical tools. First, \gls{flops} serve as a reliable early-stage predictor
of inference latency, allowing for hardware-aware model selection without
exhaustive testing. Second, the Pareto front, which visually maps the
energy–accuracy trade-offs per cycle time, guiding developers toward optimal
processor-model pairings.

\subsection{Limitations}
Despite its practical insights, this study has several limitations. Power measurements
were conducted under controlled laboratory conditions and may
not fully capture variability introduced by environmental factors, peripheral
usage, or long-term deployment effects. In addition, the analysis focuses
on a limited set of microcontroller architectures, which may restrict the
generalizability of the conclusions to newer or heterogeneous platforms.
Finally, while \gls{flops} correlate well with inference latency in the evaluated
setups, this relationship may weaken for models with complex memory access
patterns or hardware-specific accelerations.

\subsection{Future Work}
Future research should focus on refining power-measurement methodologies,
particularly through standardized evaluations of idle current and wake-up
latency to enable fair cross-processor comparisons. Furthermore, long-term,
real-world deployments are required to validate the robustness of these
findings under environmental variation, workload drift, and hardware aging.
Extending the framework to additional architectures and accelerator-based
systems would further strengthen its applicability.

\section{Conclusion}
\label{sec:conclusion}
This work presented a framework for benchmarking \gls{ai} models on baremetal
ARM Cortex processors, revealing that the optimal system design
is dictated by the application’s operational cycle. The core finding is that
the balance between active computation energy and passive idle energy not
only determines the best processor choice but also shapes the ideal model
optimization strategy.

For frequent-inference tasks, the Cortex-M7 is superior, and model efficiency
is critical for minimizing energy use. For tasks with long idle periods,
the Cortex-M4 is the clear winner, allowing developers to prioritize model
accuracy over inference speed with minimal energy penalty. Our framework,
which uses \gls{flops} as a latency predictor and Pareto analysis to visualize
trade-offs, provides a structured basis for navigating these decisions.
Ultimately, these findings guide developers toward creating embedded \gls{ai}
applications that are both sustainable and high performing by aligning hardware
selection and model optimization with the specific demands of the
real-world use case.

\section*{Acknowledgment}
This work is part of the GreenICT@FMD project and is funded by the
German Federal Ministry for Research, Technology and Space (BMFTR)
(grant number 16ME0491K).


\bibliographystyle{IEEEtran}
\bibliography{references}

\end{document}